# AN SVM MULTICLASSIFIER APPROACH TO LAND COVER MAPPING

**Gidudu Anthony**, Post Doctoral Fellow
**Hulley Gregg,** Masters Student
**Marwala Tshilidzi**, Professor
School of Electrical and Information Engineering
University of the Witwatersrand
Johannesburg, Private Bag X3, Wits, 2050, South Africa
Anthony.Gidudu@wits.ac.za, greghul@icon.co.za, Tshilidzi.Marwala@wits.ac.za

## ABSTRACT

From the advent of the application of satellite imagery to land cover mapping, one of the growing areas of research interest has been in the area of image classification. Image classifiers are algorithms used to extract land cover information from satellite imagery. Most of the initial research has focussed on the development and application of algorithms to better existing and emerging classifiers. In this paper, a paradigm shift is proposed whereby a 'committee' of classifiers is used to determine the final classification output. Two of the key components of an ensemble system are that there should be diversity among the classifiers and that there should be a mechanism through which the results are combined. In this paper, the members of the ensemble system include: Linear SVM, Gaussian (Radial Basis Function) SVM and Quadratic SVM. The final output was determined through a simple majority vote of the individual classifiers. From the results obtained it was observed that the final derived map generated by an ensemble system can potentially improve on the results derived from the individual classifiers making up the ensemble system. The ensemble system classification accuracy was, in this case, better than the linear and quadratic SVM result. It was however less than that of the RBF SVM. Areas for further research could focus on improving the diversity of the ensemble system used in this research.

**Key Words:** Ensemble Systems, Support Vector Machines, Land Cover Mapping

## INTRODUCTION

One of the means through which land cover classes can be extracted from satellite imagery is by the use of algorithms called image classifiers. Image classification may be categorized into supervised or unsupervised, parametric or nonparametric, contextual or noncontextual classification (Keuchela et al, 2003). This paper explores the use of nonparametric supervised classification algorithms called Support Vector Machines (SVMs). SVMs are nonparametric in the sense that they do not attempt to model the distribution of the training data, but try to separate the different classes by directly searching for adequate boundaries between them (Keuchel, 2003). This is unlike traditional classifiers such as maximum likelihood and minimum-distance-to-means classifiers which fall under the category of parametric classifiers. The interest in the exploration of new and emerging classifiers stems from the importance of land cover information to various disciplines such as forestry, precision agriculture, disaster management etc. How accurate a land cover map is derived has implications on how well the various application areas will be effected, be it at policy or operational level. Most of the current research in the area tends to focus on the application of new algorithms to land cover mapping with an emphasis on how they compare or if they are better than existing methods. In this paper a paradigm shift is proposed whereby instead of looking at which classifier is better than the traditional and/or emerging classifiers, there is an interest in how the classifiers can be considered collectively. The final land cover class assigned to a pixel is dependent on a vote between the 'committee' of classifiers. Hence the name – ensemble classifiers. This paper gives an overview on SVMs which are the subject of this paper. The paper then continues to highlight the issues pertaining to ensemble classification, the developed methodology and the results thereof.

# SUPPORT VECTOR MACHINES

SVMs, like other nonparametric classifiers such as Artificial Neural Networks, boast a robustness that has spearheaded its application into many areas. Having started off as a Statistical Learning Theory (Vapnik, 1995), SVMs have continued to be used in machine vision fields such as character, handwriting digit and text recognition (Vapnik, 1995; Joachims, 1998). More recently, their application to land cover mapping has been vigorously explored (Huang et al, 2002; Mahesh and Mather, 2003, Gidudu et al, 2007). Like other supervised classifiers, training data is a prerequisite to define the decision boundaries within the feature space, based upon which classification decision rules are made. For SVMs, this decision boundary is a linear discriminant placed midway between the classes of interest. Unfortunately, land cover classes when projected to the input space are rarely linearly separable. SVMs handle such datasets by nonlinearly projecting the training data in the input space to a feature space of higher (infinite) dimension by use of a kernel function. This results in the previously nonlinear datasets becoming linearly separable. Placing a linear discriminant in this high (infinite) dimension will be equivalent to placing a non linear discriminant in the previous input space. Some examples of functions (also called kernels) used to this effect include: polynomial, gaussian (more commonly referred to as radial basis functions) and sigmoid functions. Each function has parameters which have to be determined prior to classification and they are usually determined through a cross validation process. Operating in high dimension potentially renders the risk of overfitting in the input space possible. SVMs control this through the principle of Structural Risk Minimization (Vapnik, 1995). The empirical risk of misclassification is controlled by maximizing the margin between the training data and the decision boundary (Mashao, 2004). In practice this criterion is softened to the minimization of a cost factor involving both the complexity of the classifier and the degree to which marginal points are misclassified, and the tradeoff between these factors is managed through a margin of error parameter (usually designated C). Like the respective function parameters, this C parameter is tuned through cross-validation procedures (Mashao, 2004). Some of the classical literature relating to SVMs can be found in Vapnik (1995), Campbell (2000) and Christianini (2002).

# ENSEMBLE CLASSIFIERS

Ensemble systems come under different names such as multiple classifier systems, committee of classifiers or mixture of experts. The idea behind ensemble systems is to have the final classification result dependant on a pool of classifiers. Of importance to the generation of an ensemble system is that each individual classifier must be unique in how it generates decision boundaries (Polikar, 2006). The term used in ensemble systems is that there must be diversity in the ensemble system. The rational behind ensuring diversity is that each classifier will make a different error, and strategically combining these classifiers can reduce the total error (Parikh and Polikar, 2007). Some of the ways through which diversity can be ensured include: using different datasets to train different classifiers, using different training parameters for different classifiers, using different types of classifiers and using different features for the different classifiers (Polikar, 2006). Another key aspect about ensemble systems is how to combine the results from the individual classifiers. Two examples of common combination rules include simple majority and weighted majority A more theoretical treatise on ensemble systems can be found in Polikar (2006). Figure 1 gives a graphical illustration of an ensemble system.

# METHODOLOGY

The study area for this research was extracted from a 2001 Landsat scene (row 171 and row 60). It refers to Kampala which is the capital of Uganda. The land cover classes of interest were water, built up areas, think swamps, light swamps and other vegetation. IDRISI Andes was used for the preliminary processing such as identification of training samples. These data were then exported to MATLAB (version 7) for SVM classification. The content of the ensemble system was: linear, RBF and a quadratic SVM. Given that the decision boundaries for these classifiers are different, it was assumed they would provide the requisite diversity. The final classification output was determined through a majority vote of the individual classifiers. The classification results were imported into IDRISI for georeferencing, GIS integration, derivation of land cover maps and accuracy assessment.

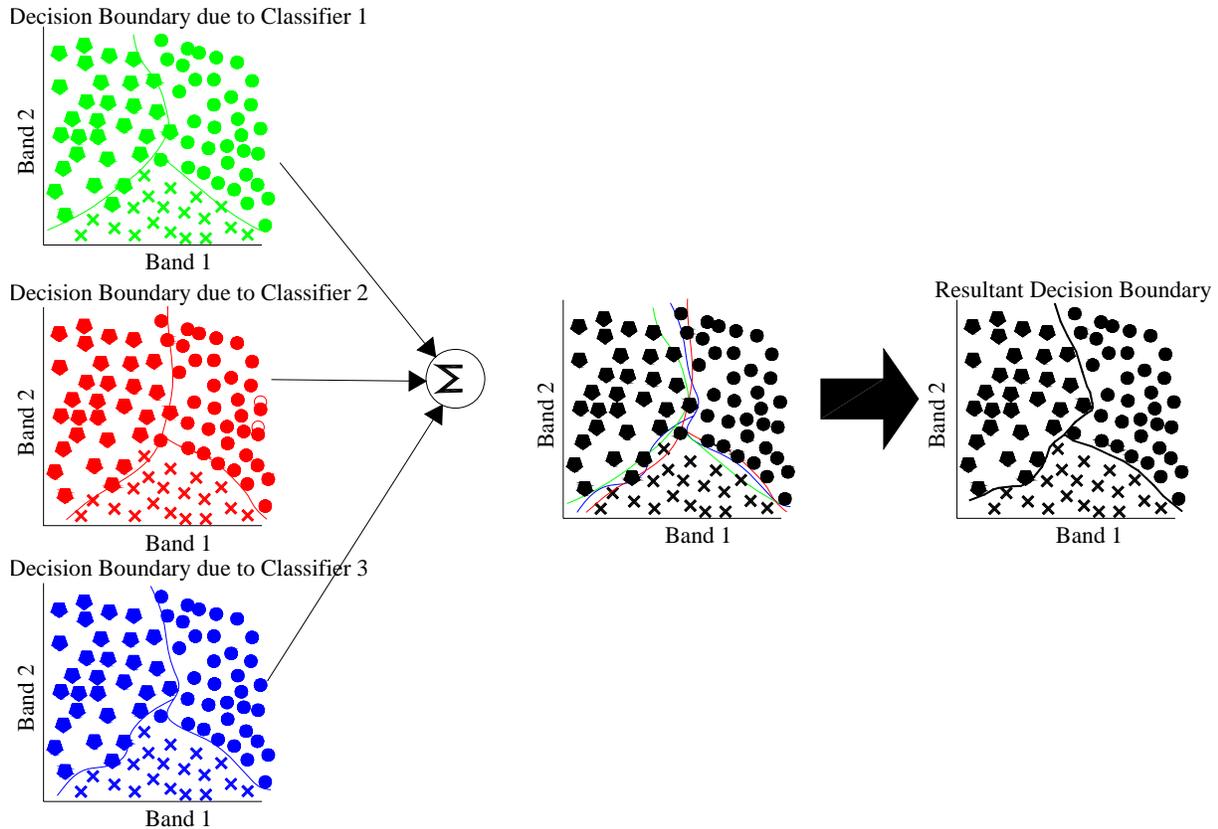

**Figure 1.** Graphical illustration of an Ensemble classifier system (Adopted from Parikh and Polikar, 2007).

## RESULTS, DISCUSSIONS AND CONCLUSIONS

In the presentation of these results, comparison is made between derived maps from the individual SVMs and the ensemble system. Figure 2 shows the derived map following the utilisation of the linear classifier. Similar land cover maps were derived for the RBF and quadratic SVMs and are illustrated in Figures 3 and 4 respectively. The key observation in this map is the presence of mixed pixels which are characteristic of linear SVM classifiers. The RBF and quadratic classifiers are less prone to mixed pixels. Figure 5 shows the result of ensemble system. From the derived maps it can be seen that the ensemble system has greatly enhanced the visual appeal of the linear SVM. Secondly, from a visual perspective, the derived map from the ensemble system is similar to that of the rbf and quadratic SVM. To further appreciate these results, error matrices were derived for each of these maps from which the overall kappa was calculated. A summary of the results is shown in Table 1.

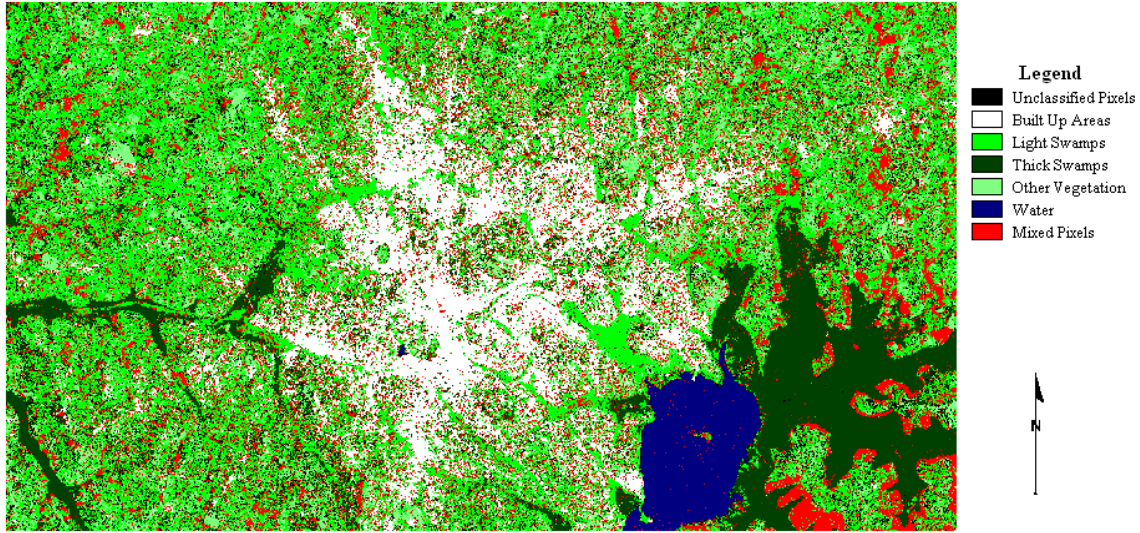

**Figure 2.** Linear SVM Classification results.

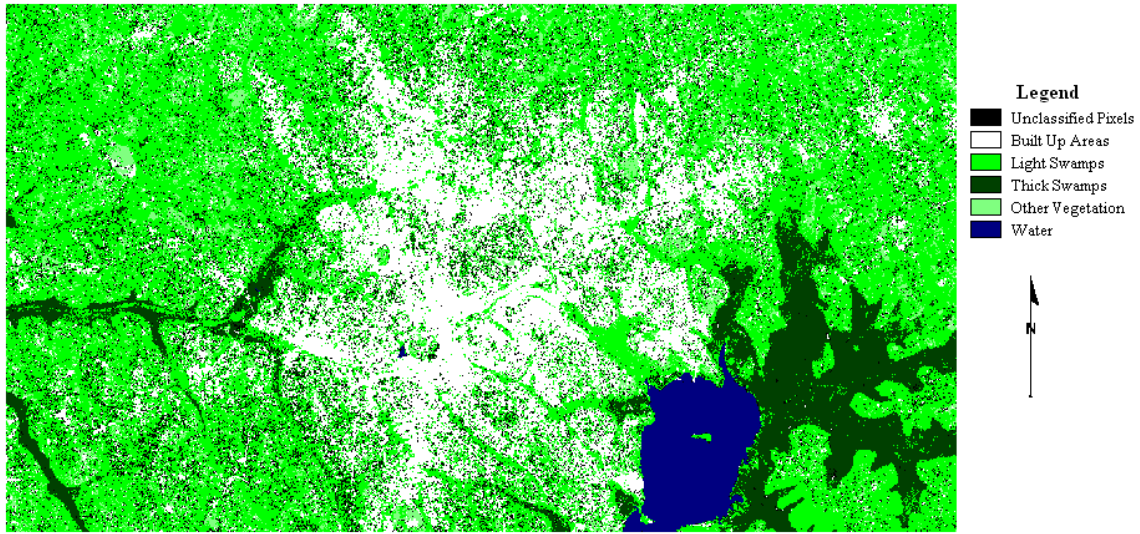

**Figure 3.** RBF SVM Classification results.

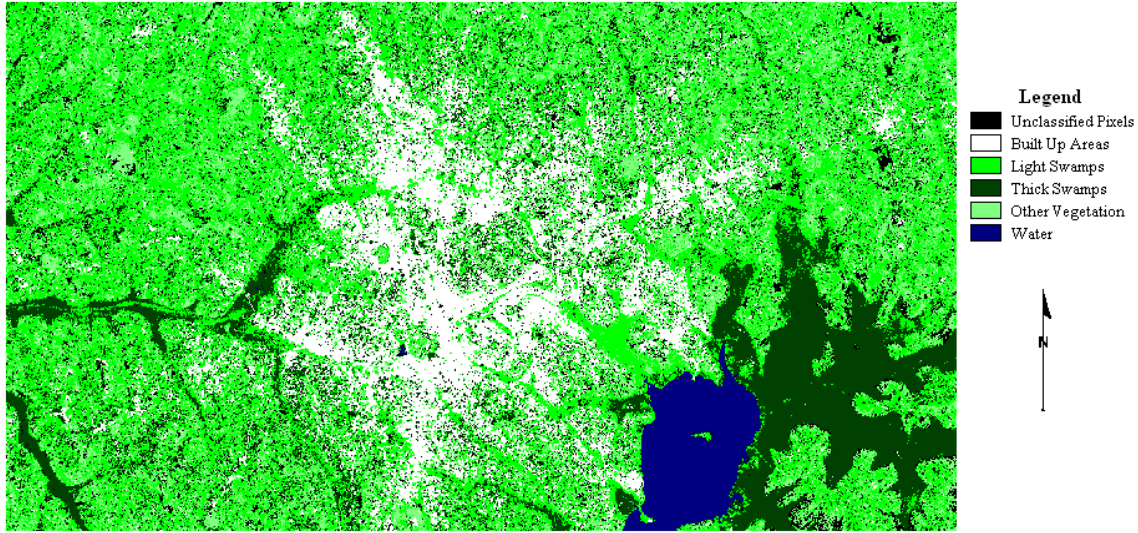

**Figure 4.** Quadratic SVM Classifier.

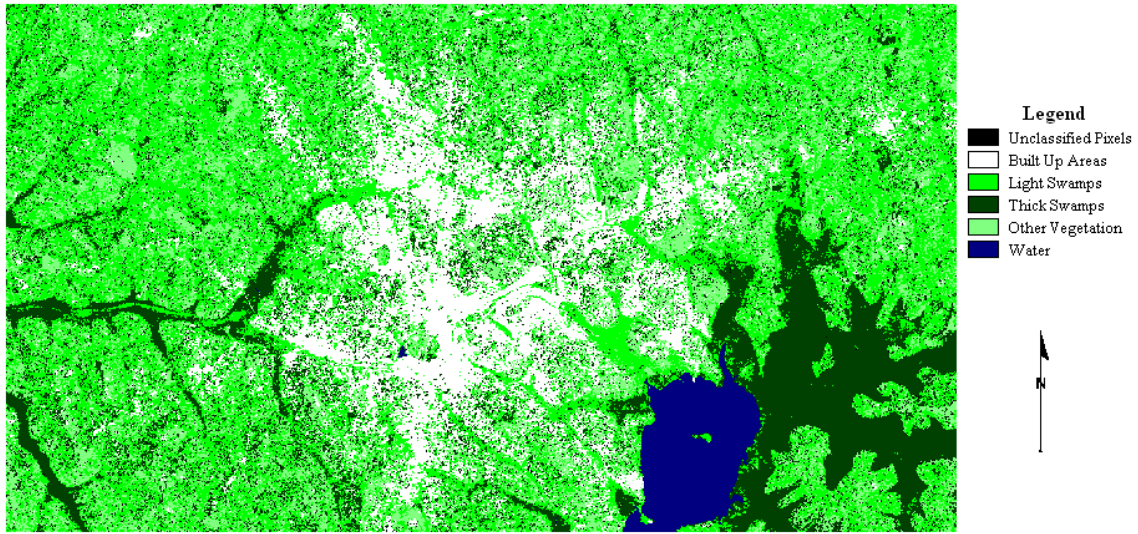

**Figure 5.** SVM Ensemble Classification.

**Table 1.** Comparison of classifiers

| SVM Classifier | Kappa |
|---|---|
| Linear | 0.7806 |
| RBF | 0.9198 |
| Quadratic | 0.8378 |
| Ensemble | 0.8929 |

From the presented results, it can be seen that the ensemble system has posted better accuracy assessment results compared to the linear and quadratic SVMs. It however has posted inferior accuracy assessment results compared to the RBF SVM. The results show that the ensemble system could potentially improve on the results of some of the individual classifiers. Polikar (2006) posits that whereas the ensemble system results may not beat the performance of the best classifier in the ensemble, it certainly reduces the overall risk of making a poor classifier selection. This ensemble classifier can certainly be improved by increasing its diversity.

# REFERENCES

Christianini, N., and Shawe-Taylor, J., 2000. *An Introduction to Support Vector Machines: And Other Kernel-based Learning Methods*, Cambridge and New York: Cambridge University Press.
Gidudu, A., Hulley, G., and Marwala, T., 2007. Image classification using SVMs: One-against-one vs one-against-all, In *Proceedings of the 28$^{th}$ Asian Conference of Remote Sensing, Kuala Lumpur, Malaysia,* 12$^{th}$ – 16$^{th}$ November 2007.
Huang, C., Davis, L. S., and Townshed, J. R. G., 2002. An assessment of support vector machines for land cover classification, *International Journal of Remote Sensing*, 23, 725–749.
Joachims, T., 1998. Text categorization with support vector machines—learning with many relevant features, In *Proceedings of the 10$^{th}$ European Conference on Machine Learning, Chemnitz, Germany,* Berlin: Springer, pp. 137–142.
Keuchela, J., Naumanna, S., Heilera, M., and Siegmund, A., 2003. Automatic land cover analysis for Tenerife by supervised classification using remotely sensed data,. *Remote Sensing of Environment,* 86, 530–541.
Mahesh P., and Mather, P. M., 2003. Support vector classifiers for land cover classification, In *Proceedings of the 6$^{th}$ Annual International Conference, Map India 2003, New Delhi, India*, 28$^{th}$ – 31$^{st}$ January 2003.
Mashao D., 2004. Comparing SVM and GMM on parametric feature-sets, In *Proceedings of the 15$^{th}$ Annual Symposium of the Pattern Recognition Association of South Africa, Cape Town, South Africa*, 27$^{th}$ – 29$^{th}$ November 2004.
Parikh, D. and Polikar, R., 2007. An ensemble-based incremental learning approach to data fusion, *IEEE Transactions on Systems, Man, and Cybernetics—Part B: Cybernetics*, Vol. 37, No. 2, April 2007.
Polikar, R., 2006. Ensemble based systems in decision making, *IEEE Circuits and Systems Magazine*, pp 21 - 44
Vapnik, V. N., 1995. *The Nature of Statistical Learning Theory*, New York: Springer-Verlag.